# Semi-Supervised Defect Detection via Conditional Diffusion and CLIP-Guided Noise Filtering

**Shuai Li[1,&], Shihan Chen[2,&], Wanru Geng[3,&], Zhaohua Xu[4], Xiaolu Liu[5], Can Dong[6], Zhen Tian[7], Changlin Chen[8,*]**

[1] School of Materials Science and Engineering，Hefei University of Technology; Hefei, 230009, China; 2022216057@mail.hfut.edu.cn

[2] College of Food Science and Technology, Jiangnan University; Jiangsu, 212013, China; csh2412586479@outlook.com

[3] School of Mechanical Engineering, Hefei University of Technology; Hefei, 230009, China; 2022211618@mail.hfut.edu.cn

[4] Electrical Engineering and Its Automation, Hefei University of Technology; Hefei, 230009, China; 15375919939@163.com

[5] College of Civil and Hydraulic Engineering, Hefei University of Technology; Hefei, 230009, China; Lau3501825750@outlook.com

[6] School of Mathematics, Hefei University of Technology; Hefei, 230009, China; dc2417353092@163.com

[7] James Watt School of Engineering, University of Glasgow , G128QQ, UK, 2620920Z@student.gla.ac.uk

[8] Institute of Humanoid Robots, Department of Precision Machinery and Precision Instrumentation, University of Science and Technology of China, Hefei 230026, China; cL--c@outlook.com

[&] These authors have made the same contribution and are co-first authors.

[*] Corresponding author.

**Abstract:** In the realm of industrial quality inspection, defect detection stands as a critical component, particularly in high-precision, safety-critical sectors such as automotive components, aerospace, and medical devices. Traditional methods, reliant on manual inspection or early image processing algorithms, suffer from inefficiencies, high costs, and limited robustness. This paper introduces a semi-supervised defect detection framework based on conditional diffusion (DSYM), leveraging a two-stage collaborative training mechanism and a staged joint optimization strategy. The framework utilizes labeled data for initial training and subsequently incorporates unlabeled data through the generation of pseudo-labels. A conditional diffusion model synthesizes multi-scale pseudo-defect samples, while a CLIP cross-modal feature-based noise filtering mechanism mitigates label contamination. Experimental results on the NEU-DET dataset demonstrate a 78.4% mAP@0.5 with the same amount of labeled data as traditional supervised methods, and 75.1% mAP@0.5 with only 40% of the labeled data required by the original supervised model, showcasing significant advantages in data efficiency. This research provides a high-precision, low-labeling-dependent solution for defect detection in industrial quality inspection scenarios. The work of this article has been open-sourced at https://github.com/cLin-c/Semisupervised-DSYM

## 1. Introduction

In the Industry 4.0 era, stringent quality requirements have made defect detection a critical component of modern manufacturing and engineering systems. Statistics show that high-quality defect detection systems can reduce product recall rates by 30–50%, significantly lowering businesses' economic losses. In high-precision, safety-critical fields such as automotive components, aerospace, and medical devices, minor defects are not just quality issues, but also potential risks that threaten life safety.

Currently, defect detection often relies on experienced technicians conducting direct (visual) or indirect (using optical instruments, etc.) inspections[1]. However, this traditional method has drawbacks, including low efficiency, high costs, and stringent regulatory requirements[2]. Early defect detection primarily used traditional image processing algorithms, such as threshold segmentation, edge detection, and template matching. These methods require manual extraction of feature rules and are highly sensitive to changes in lighting and background interference, which makes it difficult to detect minor defects. In dynamic production line environments, a lack of robust algorithms may lead to an increased number of false positives and false negatives.

Breakthroughs in deep learning and high-resolution imaging technology in recent years have provided new opportunities for defect detection. Learning algorithms based on convolutional neural networks automatically identify defect features using large amounts of data and perform end-to-end training on small samples. These algorithms can significantly improve the accuracy and robustness of defect detection by iteratively optimizing the training results on their own[3]. The development of society and the market has driven the demand for real-time online detection, prompting researchers to combine advanced technologies, such as transfer learning[4], generative adversarial networks[5], and few-shot learning, with traditional machine vision. This combination addresses new challenges, such as data scarcity, complex background interference, and uneven defect category distribution.

Nowadays, there are two main paradigms in learning algorithms. The first is supervised learning models, which establish a mapping relationship between input features and output labels by minimizing the difference between predicted and actual values. The second paradigm is unsupervised learning models, which automatically discover patterns in data distributions by measuring the similarity or differences between data points. However, supervised learning models require a large amount of labeled data, which is difficult to obtain in real industrial production environments. Unsupervised learning models often fail to meet industrial accuracy requirements. Achieving a defect detection system with higher accuracy, efficiency, and generalizability remains a critical technical challenge for academia and industry. To adapt to situations with limited labeled data and abundant unlabeled data in real-world industrial scenarios, this paper proposes a semi-supervised detection framework based on conditional diffusion. This framework reduces reliance on labeled data, enhances model robustness, addresses class imbalance issues, and balances computational efficiency and scalability while adapting to dynamic data environments. The framework is divided into two parts: supervised and unsupervised training. An adaptive conditional diffusion method generates defect data from original defective images, which are then used for supervised training alongside original images. Since suspicious areas are difficult to distinguish, noisy labels may be generated. Therefore, the CILP model is used for pre-filtering obviously defect-free images in the unsupervised part to prevent incorrect pseudo-labels from contaminating the training and to achieve noise processing. The method presented in this paper achieved an mAP@0.5 of 78.4% on the NEU-DET dataset, which represents a significant improvement over existing semi-supervised methods. This validates the effectiveness of the diffusion-guided teacher-student framework in industrial defect detection. The main contributions of this paper are as follows:

1. A semi-supervised detection framework based on conditional diffusion was designed that fully utilizes labeled and unlabeled samples, thereby reducing the amount of data annotation required.
2. A detection head based on Mamba was designed to improve the accuracy of labeled training and generate higher-precision pseudo labels for unsupervised training.
3. A noise pseudo-label filtering mechanism based on CLIP was constructed to mitigate pseudo-label error issues in semi-supervised learning.

## 2.Related Works

Industrial defect detection, as an important research direction in the fields of computer vision and artificial intelligence, has made remarkable progress in recent years driven by deep learning technology. This section sorts out the current research status from the perspectives of deep learning detection methods, semi-supervised learning techniques, applications of generative models, and multimodal fusion, providing a theoretical basis for the methods proposed in this paper.

*2.1 Defect detection method based on deep learning*

The rapid development of deep learning technology has brought revolutionary changes to industrial defect detection. The early convolutional Neural network (CNN) methods were mainly based on the classification idea and identified defect types by extracting the hierarchical features of images. Although the LeNet network proposed by LeCun et al. [6] was initially used for handwritten digit recognition, its convolution operation and pooling mechanism laid the foundation for subsequent defect detection. The success of AlexNet[7] further proves the advantages of deep CNN in image recognition tasks, prompting researchers to apply it to industrial quality inspection scenarios.

With the continuous optimization of the network structure, ResNet[8] solved the vanishing gradient problem in the training of deep networks by introducing residual connections, enabling the network to effectively learn deeper feature representations. Zhang et al. proposed a surface defect detection method based on multi-scale feature fusion based on the ResNet architecture, and improved the detection accuracy by fusing feature information at different levels. DenseNet[9] further enhances feature reuse through dense connections and performs well in defect detection of small targets. Chen et al. Proposed an unsupervised anomaly detection method based on a dual-tower reconstruction network. Through local and global feature denoising and reconstruction as well as selective fusion modules, the accuracy of industrial defect location was significantly improved[10].

The development of the object detection framework provides new solutions for defect detection. Two-stage detectors such as Faster R-CNN[11] achieve end-to-end object detection through the design of region Proposal Network (RPN) and classification regression heads. The YOLOv series [12-14], as a single-stage detector, significantly enhances the inference speed while ensuring detection accuracy, making it more suitable for real-time industrial detection requirements.

*2.2 The Application of Semi-supervised Learning in Defect Detection*

The problem of scarce labeled data in industrial scenarios has given rise to the development of semi-supervised learning methods. Semi-supervised learning effectively alleviates the problem of high data annotation costs by training the model simultaneously using a small amount of labeled data and a large amount of unlabeled data.

The teacher-student framework is one of the mainstream paradigms of semi-supervised learning. The Mean Teacher method proposed by Tarvainen and Valpola[15] updates the parameters of the teacher model through exponential moving average (EMA), and uses the teacher model to generate pseudo-labels for unlabeled data to guide the learning of the student model. This method has achieved excellent results in the image classification task and has subsequently been widely applied to the object detection task. The FixMatch proposed by Sohn et al. [16] further improves the quality of pseudo-labels through consistency constraints enhanced by strong and weak data.

In the field of defect detection, Xu et al. [17] combined self-training and cooperative training strategies and achieved remarkable results in the detection of steel surface defects. However, the existing methods still have deficiencies in dealing with pseudo-label noise, especially in complex industrial environments, where background interference is prone to be misidentified as defects.

Consistency regularization is another important semi-supervised learning strategy. This method is based on the assumption that different perturbation versions of the same input should produce consistent predictions. Π-Model[18] and Temporal Ensembling learn stable feature representations by minimizing the prediction differences at different times or under different disturbances. The application of these methods in defect detection mainly focuses on improving the robustness of the model to environmental changes such as illumination and Angle.

*2.3 The Application of Generative Models in Defect Detection*

Data imbalance and sample scarcity are the core challenges faced by industrial defect detection. Generative models synthesize new samples by learning the data distribution, providing an effective way to solve this problem.

Generative Adversarial Network (GAN) has been widely applied in defect data enhancement. The original GAN proposed by Goodfellow et al. learns the data distribution through adversarial training of the generator and discriminator. In the field of defect detection. The success of DCGAN[19] has further promoted the application of GAN in image generation tasks. Lu et al. Designed a lightweight generative adversarial network. Through inception block replacement and residual optimization, the number of parameters was reduced by 38% to 961k, while achieving 27.41 PSNR on the Urban100 dataset, providing an efficient solution for industrial image super-resolution[20].

However, GAN training has problems such as pattern collapse and unstable training, which limit its effect in defect data generation. To solve these problems, researchers have proposed a variety of improvement plans. WGAN[21] improved the training stability through Wasserstein distance, and Progressive GAN[22] enhanced the generation quality through progressive training. As an emerging generative model, the diffusion model shows significant advantages in terms of image generation quality and training stability. The DDPM proposed by Ho et al. [23] achieves high-quality image generation through the denoising and diffusion process. Dhariwal and Nichol[2] further demonstrated that the diffusion model outperforms GAN in the task of image synthesis. In defect detection applications, the diffusion model shows unique advantages. Yang et al. [24] utilized the conditional diffusion model to generate defect samples of specific types, significantly improving the detection performance of rare defect categories. Compared with GAN, the defect images generated by the diffusion model have higher quality and a more stable training process, but the reasoning speed is relatively slow.

## 3. Methods

The present paper proposes a diffusion-guided semi-supervised defect detection framework (Diffusion-guided Semi-supervised YOLOv9-Mamba, DSYM) that addresses the key issues of scarce labeled data and pseudo-label noise interference in industrial defect detection. This framework achieves efficient integration and utilization of limited labeled data and large amounts of unlabeled data through multimodal feature fusion and a staged collaborative optimization strategy. This section will delineate the core components and technical implementation of the framework.

The fundamental design philosophy of the DSYM framework is to construct an end-to-end, multi-stage training pipeline. The training process employs a two-stage strategy: in the first stage, labeled data is combined with high-quality defect samples generated by conditional diffusion for supervised pre-training to establish reliable basic detection capabilities; in the second stage, a large amount of unlabeled data is introduced, and knowledge distillation and model performance improvement are achieved through the collaborative optimization of teacher-student networks and intelligent noise filtering of the CLIP model.

*3.1 Semi-supervised training framework*

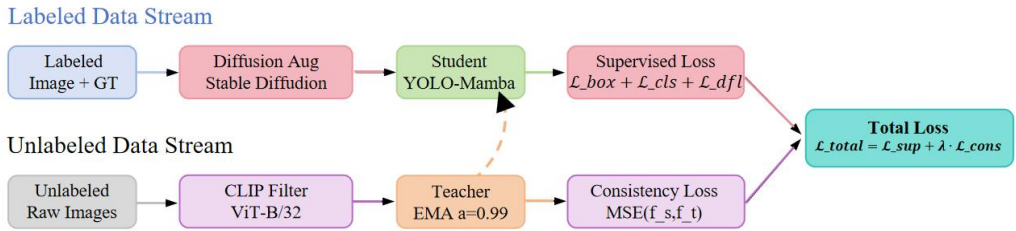

**Figure 1.** Semi-supervised training framework

The semi-supervised learning framework designed in this paper uses a teacher-student network collaboration mechanism to train the model. First, the student model undergoes initial supervised training using a small amount of labeled data. Then, the teacher model is initialized using the student model to generate pseudo labels and guide its further learning. The steps for generating pseudo labels include:

a. The teacher model infers unlabeled data to generate initial pseudo labels.
b. The CLIP model filters out noisy labels, retaining only high-confidence defect regions.
c. The student model is trained through joint supervision with high-quality pseudo and real labels to optimize its feature expression capabilities.

The term "training loss" encompasses two distinct components: supervised loss and consistency loss. The precise definitions of these components are as follows:
The term "supervised loss" is defined as follows:

$$l_{sup} = \frac{1}{N}\sum_{i=1}^{N} CrossEntropy(p_i, t_i) \tag{1}$$

The term "consistency loss" has been employed to denote the process of ensuring consistency between the outputs of the teacher and student models, as delineated by the mean square error (MSE).

$$L_{consistency} = \frac{1}{N}\sum_{i=1}^{N} ||s_i - t_i||^2 \tag{2}$$

In this system, pi represents the prediction of the student model, ti represents the true or teacher pseudo label, and si and ti represent the outputs of the student and teacher models, respectively.

The teacher model parameter update employs the exponential moving average (EMA) strategy.

$$\theta_t \leftarrow \alpha\theta_t + (1-\alpha)\theta_s \tag{3}$$

The student model employs standard backpropagation methods to update parameters, thereby enhancing its learning flexibility.

*3.2 The Design of The Probe*

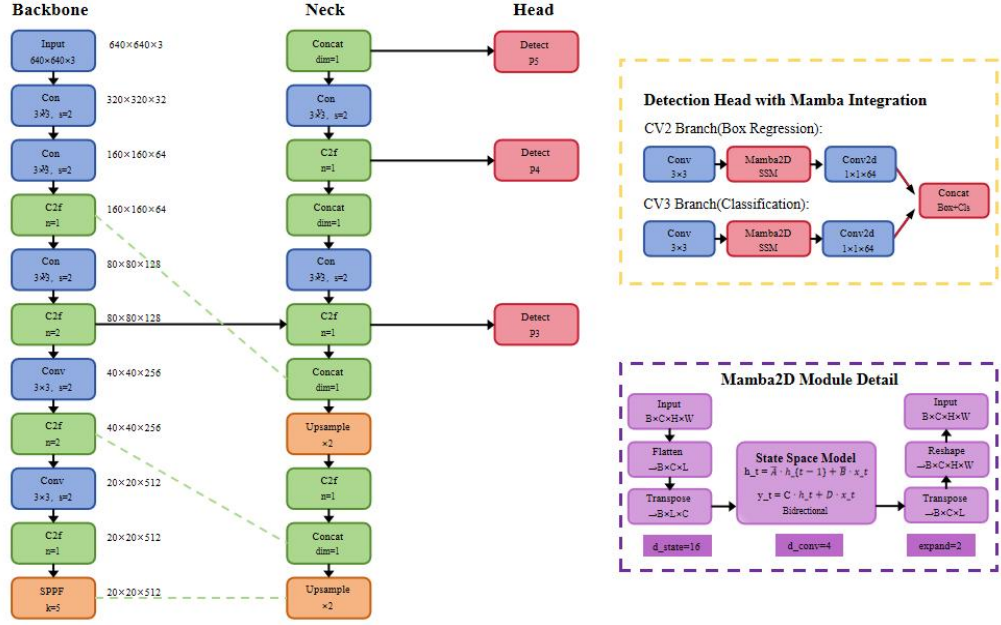

**Figure 2.** Detection head structure

In order to enhance the model's capacity to extract defect features at multiple scales, this paper proposes the integration of the Mamba state space model into the YOLOv9 detection head. The Mamba model has been demonstrated to efficiently process long sequence dependencies through a selective state space mechanism, rendering it particularly suitable for handling the complex spatial distribution patterns of industrial defects. The Mamba model utilizes the following state space equation to perform feature transformation for the input feature sequence f{$x$} = ($x\_1, x\_2, ..., x\_L$):

$$h_t = Ah_{t-1} + Bx_t$$
$$y_t = Ch_t + Dx_t \qquad (4)$$

Among these, A ∈ $R^{(N×N)}$ is the state transition matrix, B ∈ $R^{(N×D)}$ and C ∈ $R^{(D×N)}$ are the input and output projection matrices, respectively, and D ∈ $R^{(D×D)}$ is the skip connection matrix.

The detection head designed in this paper adopts a decoupled structure, separating classification and regression tasks to improve detection accuracy. The detection head comprises three branches: the bounding box regression branch, the classification branch, and the confidence branch. The following investigation focuses on the bounding box regression branch. The distributed focus loss (DFL) mechanism is employed for bounding box prediction, thereby transforming the regression problem into a classification problem.

$$B = f(d, a, s) \qquad (5)$$

Among them, *d* is the distribution prediction output by the DFL layer, *a* is the anchor box coordinates, *s* is the current layer stride, and *f* is the coordinate transformation function.

Given the heterogeneity and imbalance of defect categories, an enhanced Sigmoid activation function is employed:

$$P(c) = \sigma(c) = \frac{1}{1+e^{-c}} \qquad (6)$$

*c* is the category prediction element of the model output.

To expedite the convergence of models, an adaptive initialization of the detection head bias is implemented.

$$b = \ln\left(\frac{n}{s}\right) \qquad (7)$$

Among them, *n* denotes the number of categories, and *s* indicates the stride of the current detection layer.

*3.3 Condition diffusion model data augmentation*

In order to address the scarcity of defect samples and class imbalance, this paper proposes a conditional diffusion model to generate high-quality synthetic defect samples. The model generates realistic defect images based on defect class information and spatial location information. The forward diffusion process is defined as follows:

$$q(x_{1:T}|x_0) = \prod_{t=1}^{T} q(x_t|x_{t-1}) \tag{8}$$

The transition probability of each step is as follows:

$$q(x_t|x_{t-1}) = N(x_t; \sqrt{1-\beta_t}x_{t-1}, \beta_t I) \tag{9}$$

Among them, $\beta_t$ is a predefined noise scheduling parameter. The reverse denoising process is performed by a learned network $\sigma_\theta$.

$$p_\theta(x_{t-1}|x_t, c) = N(x_{t-1}; \mu_\theta(x_t, t, c), \sigma_t^2 I) \tag{10}$$

Among these, $c$ is classified as conditional information, encompassing defect category labels and spatial location data. To achieve precise control over the type and location of generated defects, this paper proposes a multi-modal conditional encoding mechanism. The following is a categorization of conditional encoding：

$$e_{cls} = Embedding(classid) + PE(position) \tag{11}$$

The following is a spatial condition coding：

$$e_{spa} = CNN(M_{mask}) + MLP(b_{box}) \tag{12}$$

Among them, $M_{mask}$ is the defect region mask, and $b_{box}$ is the bounding box coordinate. Condition fusion:

$$c = CrossAttention(e_{cls}, e_{spa}) + e_{cls} + e_{spa} \tag{13}$$

The training objective of the diffusion model is as follows:

$$\varsigma_{diffusion} = E_{t,x_0,\epsilon,c}[||\epsilon - \epsilon_\theta(x_t, t, c)||^2] \tag{14}$$

To enhance the quality of generation, a DDIM sampling strategy is employed for rapid inference.

$$x_{t-1} = \sqrt{\alpha_{t-1}}\left(\frac{x_t - \sqrt{\alpha_t}\sigma_\theta(x_t, t, c)}{\sqrt{\alpha_t}}\right) + \sqrt{1-\alpha_{t-1}}\sigma_\theta(x_t, t, c) \tag{15}$$

*3.4 CLIP-guided noise filtering mechanism*

In order to address the issue of pseudo-label noise in semi-supervised learning, this paper introduces the CLIP model to construct an intelligent noise filtering mechanism. CLIP establishes cross-modal feature alignment between images and text through contrastive learning, thereby enabling effective identification and filtering of low-quality pseudo-labels.

Initially, image features are extracted.

$$f_{img} = CLIP_{image}(I) \tag{16}$$

Subsequently, perform feature extraction on the text.

$$f_{text} = CLIP_{text}("A photo with [defect\_type] defect") \tag{17}$$

The calculation of similarity is as follows:

$$s(I,t) = \frac{f_{max} \cdot f_{max}}{||f_{max}||||f_{max}||} \tag{18}$$

To enhance the precision of noise filtering, an adaptive threshold mechanism was developed.

$$\tau_t = \tau_0 \cdot exp(-\frac{t}{T_{total}} \cdot \lambda) \quad (19)$$

In this formula, $\tau_0$ is the initial threshold, $t$ is the current training step, $T_{total}$ is the total training step, and $\lambda$ is the decay coefficient.

The following criteria have been established for the implementation of noise filtering:

$$keep\_sample = \begin{cases} True, & if s(I,t) > \tau_t \text{ and } C_{teacher} > \tau_{conf} \\ False, & otherwise \end{cases} \quad (20)$$

Among them, $C_{teacher}$ is the confidence level predicted by the teacher model, and $\tau_{conf}$ is the confidence threshold.

The following is the pseudocode for the DSYM training algorithm designed in this paper:

**Table 1.** The pseudo-code of the DSYM training algorithm.

| | Pseudo-code |
|---|---|
| 1 | Algorithm : DSYM Training Algorithm |
| 2 | Input: Labeled dataset D_l, unlabeled dataset D_u, hyperparameters α, λ, τ |
| 3 | Output: Optimized student model θ_student |
| 4 | //Phase 1: Supervised pre-training |
| 5 | for epoch = 1 to 50 do |
| 6 |     Generate conditional diffusion samples D_syn ← DiffusionModel(D_l) |
| 7 |     Train the student model: θ_student ← optimize(D_l ∪ D_syn, L_sup) |
| 8 | end for |
| 9 | //Phase 2: Semi-supervised collaborative training |
| 10 | Initialize the teacher model: θ_teacher ← θ_student |
| 11 | for epoch = 51 to 200 do |
| 12 |     for batch (x_u) in D_u do |
| 13 |       // CLIP noise filtering |
| 14 |       if CLIP_filter(x_u) and teacher_confidence(x_u) > τ then |
| 15 |         // Generate pseudo labels |
| 16 |         y_pseudo ← teacher_model(x_u) |
| 17 |         // Train the student model |
| 18 |         L_total ← L_sup + λ_unsup(t) * L_consistency |
| 19 |         θ_student ← optimize(L_total) |
| 20 |         // Teacher model EMA update |
| 21 |         θ_teacher ← α * θ_teacher + (1-α) * θ_student |
| 22 |       end if |
| 23 |     end for |
| 24 | end for |
| 25 | return θ_student |

## 4. Results

*4.1 Dataset*

The experiment selected the steel surface defect detection dataset [25] that had been made open-source by Song Kesheng's team at Northeastern University. In accordance with the methodologies and experimental designs delineated in this paper, the dataset was subsequently subdivided into new test sets, supervised training sets, unsupervised training sets, and validation sets. The specific divisions of the dataset are as follows:

**Table 2.** Dataset partitioning.

| Types of Test Project | Test | Train | Val |
|---|---|---|---|

|  |  | Supervision | Unsupervised |  |
|---|---|---|---|---|
| Crazing | 27 | 54 | 216 | 5 |
| Inclusion | 27 | 54 | 216 | 5 |
| Patches | 28 | 54 | 216 | 6 |
| Pitted-Surface | 29 | 54 | 216 | 7 |
| Rolled-In-scale | 29 | 54 | 216 | 7 |
| Scratches | 30 | 54 | 216 | 8 |
| Total | 170 | 324 | 1296 | 38 |

In this classification strategy, the supervised training set accounts for only 20% of the total training data (324/1620), which fully simulates the actual situation of scarce labeled data in industrial scenarios.

*4.2 Evaluation Criteria*

In general, the precision of the true positive samples (TP) is calculated to reflect the model's ability to "avoid false positives," that is, to correctly detect positive samples. The number of correctly detected positive samples (TP) and the number of incorrectly detected negative samples (FN) are calculated to reflect the model's ability to "not miss any true positives," that is, to avoid false negatives.

$$\text{Precision} = \frac{TP}{TP+FP} \quad (21)$$

$$\text{Recall} = \frac{TP}{TP+FN} \quad (22)$$

In this experiment, the optimization of precision or recall in isolation may result in misclassifications across various aspects. Additionally, precision and recall are typically negatively correlated, meaning that enhancing precision may concurrently diminish recall, and vice versa. In general, following the sorting process based on confidence level, the F1-Score is calculated by adjusting the threshold to achieve a balance between precision and recall. Subsequently, a Precision-Recall (PR) curve is plotted, and a comprehensive evaluation is conducted using MAP.

Mean Average Precision (MAP) is a core evaluation metric in object detection and information retrieval. This metric is used to measure the comprehensive performance of a model in multi-category tasks. During the calculation process, the model must sort the detection results by confidence level, adjust the threshold, plot PR curves for different categories, integrate the area under the curve (AP), and then take the average of all category AP values to obtain the MAP value.

The rigidity of the maximum a posteriori (MAP) evaluation is contingent upon the IOU threshold (the overlap between the detected boxes and the ground truth boxes). For instance, the maximum a posteriori (MAP) at 0.5 indicates an IOU threshold of 0.5, while the maximum a posteriori (MAP) at [0.5:0.95] calculates an average across multiple thresholds (COCO standard).

The AP value and mAP value (Average Precision and Mean Average Precision） are defined as follows:

$$AP = \int_0^1 P(r)dr \quad (23)$$

$$mAP = \frac{1}{C}\sum_{i=1}^{C} AP_i \quad (24)$$

Among these, average precision (AP) is represented by the shaded area enclosed by the precision-recall curves (PRCs). In this context, P(r) is a curve with recall as the horizontal axis and precision as the vertical axis. Mean average precision (mAP) is defined as the arithmetic mean of the AP values for each category. C denotes the number of defect categories in the test set. The AP (Area Under the Curve) metric is employed to assess the local performance of the learning model for each category, while the mAP (mean Average Precision) metric is used to evaluate the overall performance of the learning model across all categories [26].

*4.3 Ablation Experiment*

The present study has designed five ablation experiments on the aforementioned dataset. The baseline control group is YOLOv9. The second control group is introduced as the Mamba encoder based on the first control group. The third control group is designed as a semi-supervised detection framework based on the second control group. The fourth control group is introduced as a diffusion model based on the second control group. The results of the ablation experiments are shown in Table 3.

**Table 3.** Ablation experiment results.

| Configuration | Recall(%) | Precision(%) | mAP@0.5(%) |
|---|---|---|---|
| YOLOv9 | 72.5 | 66.1 | 74.8 |
| YOLOv9+mamba | 71.8 | 69.2 | 75.8 |
| YOLOv9+mamba+ Semi-supervised | 64.1 | 81.1 | 78.4 |
| Semi-supervised（Annotation ratio20%）+mamba | 64.5 | 58.6 | 67.5 |
| Semi-supervised（Annotation ratio 40%）+mamba | 71.2 | 68.4 | 75.1 |
| Semi-supervised（Annotation ratio 40%）+ Diffusion models | 58.2 | 63.3 | 62.1 |

The findings of the ablation experiment demonstrate that the incorporation of the Mamba encoder into the YOLOv9 model results in a 1% increase in mAP. Subsequent to the construction of a semi-supervised detection framework based on YOLOv9 and the Mamba detection head, mAP demonstrated a 3.6% improvement in comparison to the YOLOv9 model on the original identical data, reaching 78.4%.

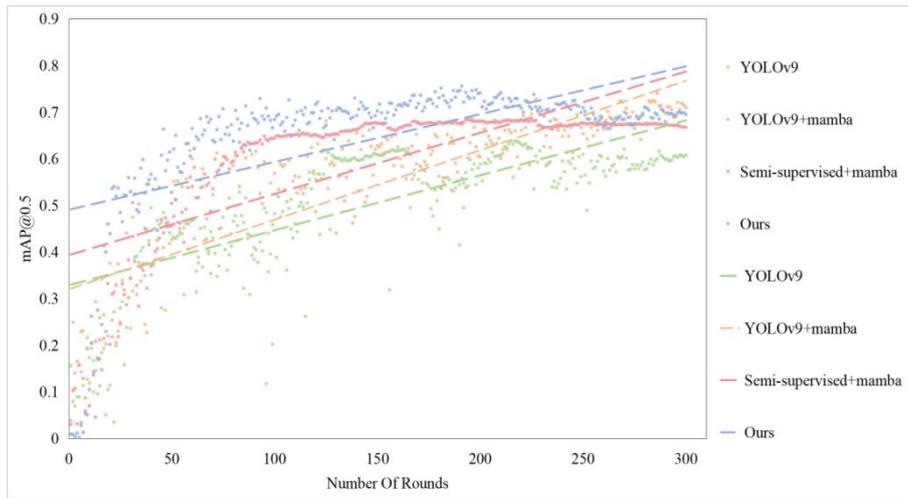

**Figure 3.** The mean amplitude (mAP) @0.5 shows an upward trend across all experiments

*4.4 Analysis and comparison of experimental results*

**Table 4.** Visualization of experimental results in different defect types.

| Defect type | Recall（%） | Precision（%） | mAP0.5（%） |
|---|---|---|---|
| Crazing | 37.5 | 64.5 | 49.3 |
| Inclusion | 59.1 | 87.3 | 80.9 |
| Patches | 93.6 | 92.4 | 97.8 |
| Pitted-Surface | 66.7 | 87.6 | 81.4 |
| Rolled-In-scale | 50.0 | 78.1 | 75.1 |
| Scratches | 77.8 | 76.6 | 86.2 |

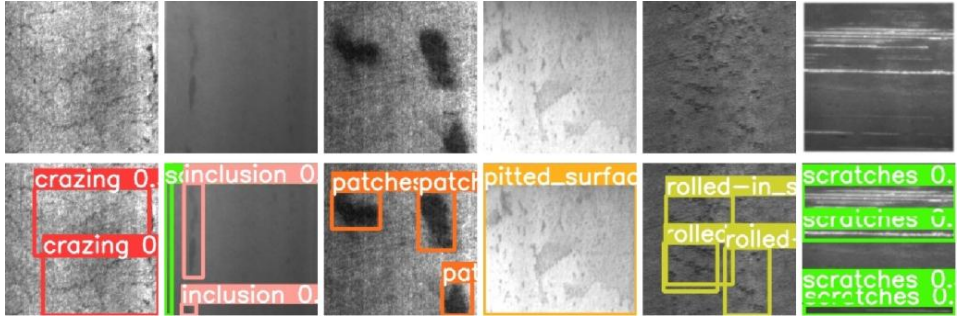

**Figure 4.** Experimental visualization results for different defect types.

This paper systematically evaluates the proposed defect detection method on six typical surface defects (cracks, inclusions, spots, pitted surfaces, rolled surfaces, and scratches). The model demonstrates a strong performance across various defect categories, with particular strengths in spot detection (mAP@0.5: 97.8%). However, there is room for enhancement in crack detection. The mAP@0.5 metric reveals that patches (97.8%), scratches (86.2%), and pitted surfaces (81.4%) exhibited the highest overall performance, attributable to their regular shapes and the availability of sufficient training samples. Conversely, the mAP@0.5 (49.3%) of opened fissures was only half that of patches, underscoring the complexity of detecting this particular defect.

The proposed methodology in this paper successfully balances high accuracy and high recall for a wide range of defect categories. Notably, it demonstrates particular efficacy in detecting common defects, such as patches and scratches.

The following text is intended to provide a comprehensive overview of the subject matter. The following table presents a performance comparison with existing methods.

**Table 5.** Performance comparison with existing methods.

| Method | Backbone Networ | mAP@0.5(%) |
|---|---|---|
| TOOD[27] | ResNet50 | 65.13 |
| YOLOv3[28] | Darknet53 | 72.3 |
| YOLOv7-tiny[29] | - | 73.7 |
| STD2[30] | Swin Large | 72.38 |
| **DSYM** | - | 78.4 |

This paper further engages in comparative analysis of methodologies employed with TOOD and STD2. As illustrated in the table, the mAP0.5 of the existing methods is lower than that obtained by the model in this paper, as indicated by a quantitative analysis. A comparison of this model with STD2 reveals an enhancement of 6.02%. The figure below illustrates the performance indicators of the model presented in this paper.

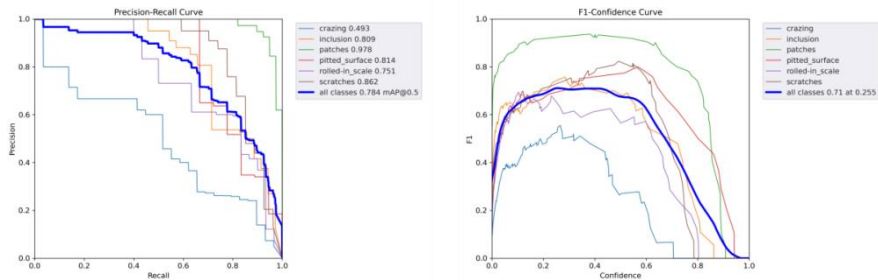

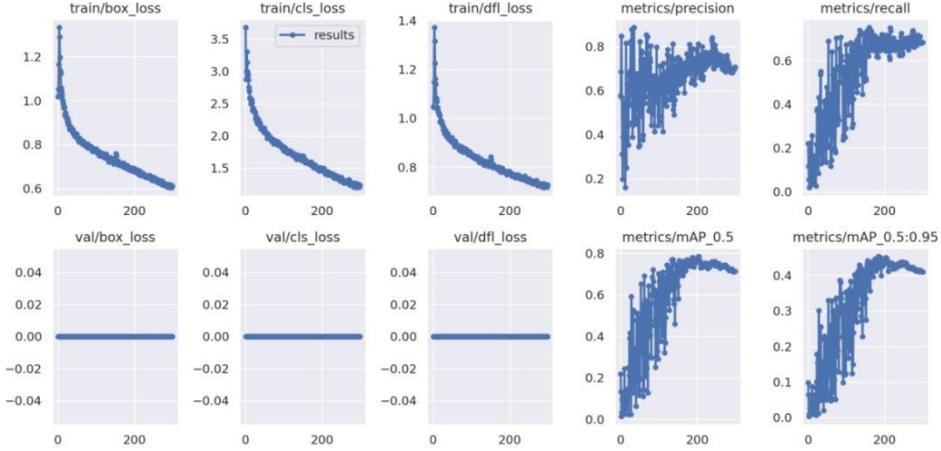

**Figure 5.** Analysis diagram of the performance index of the model

To more clearly illustrate the advantages and innovations of the method proposed in this paper, a simple comparison is made with the following five works. TMSDNet[31] proposes a 3D reconstruction framework combining multi-scale dense networks with Transformer. Support single/multiple view input. It has strong global shape perception ability and supports multi-perspective fusion. However, the computational complexity is high and the capture of details of tiny objects is insufficient. Reciprocal attention anomaly detection[32] suppresses background noise through the reciprocal attention module. It can effectively distinguish the foreground but has low sensitivity to complex texture defects (such as cracks), but there is still room for improvement in terms of computational efficiency, generalization of anomaly types, module interpretability and actual deployment. Enhanced RFCN [33] is a multi-dimensional regional convolutional network for pulmonary nodule detection, integrating multi-scale features and location-sensitive scoring maps. It has high sensitivity (98.1%) and low false positive (2.19 FP/ scan); However, it is only applicable to medical images and has limited generalization ability. EAPT[34] designs an efficient attention pyramid Transformer to balance accuracy and speed in classification/detection/segmentation tasks. It has excellent performance in multi-tasking and improved parameter efficiency. The single training configuration is complex and lacks robustness in detecting small targets. Neural foveated super-resolution for real-time VR rendering [35] is a real-time VR rendering technology based on the visual characteristics of the human eye, reducing the computational load through block super-resolution. Its edge artifacts are well controlled; However, the generation quality depends on the fixed fixation pattern and has poor dynamic adaptability. Table 6 shows a simple comparison between DSYM (ours) and the above-mentioned methods, mainly comparing multi-scale generation and the presence or absence of noise. Through the above comparison, it can be known that DSYM achieves a balance between high precision and low supervision requirements in industrial anomaly detection tasks through multi-scale feature fusion and noise suppression mechanisms.

**Tabel 6.** A comparison between DYSM and other mainstream methods

| Method | Annotation Requirement | Dataset | Multi-Scale Generation | Noise Filtering |
|---|---|---|---|---|
| TMSDNet | Full Supervision | ShapeNet/Pix3D | √ | × |
| Reciprocal Attention | Full Supervision | Medical Images | × | √ |
| Enhanced RFCN | Full Supervision | Lung Nodule Dataset | × | × |
| EAPT | Full Supervision | Custom | × | × |
| Neural Gaze SR | Semi-Supervised | Multi-Task Custom SR Dataset | √ | × |
| DSYM | 40%Supervision | NEU-DET | √ | √ |

## 5. Conclusions

The proposed methodology is a semi-supervised detection framework (DSYM) based on conditional diffusion. The proposed method establishes a two-stage collaborative training mechanism, utilizing a staged joint optimization strategy. Initially, the labeled data is employed to train the base detection model. Subsequently, unlabeled data is integrated to generate pseudo labels and facilitate training, thereby achieving efficient integration of labeled and unlabeled data. Concurrently, a conditional diffusion model is designed to generate multi-scale pseudo-defect samples, combined with CLIP cross-modal features to establish an intelligent noise filtering mechanism, effectively mitigating the label contamination issue in semi-supervised learning.

While the proposed method has yielded specific outcomes in experimental settings, it is important to acknowledge its inherent limitations. First, during the processing of unlabeled data, although the CLIP model is introduced to filter out images that are "obviously not defects," the judgment criteria of the CLIP model may have some subjectivity. For certain blurry or difficult-to-distinguish images, its filtering effect may not be ideal. Furthermore, the parameter update strategy employed in the teacher model of the teacher-student framework (e.g.EMA) may necessitate additional optimization to achieve a more balanced equilibrium between the stability of the teacher model and the adaptability of the student model.

To address these limitations, future research can be conducted in the following areas.
1. An exploration of more effective noise filtering methods is warranted. In addition to the CLIP model, other visual models or feature extraction methods can be considered to more accurately determine the presence of defects in images, thereby reducing the generation of false labels.
2. A more in-depth exploration of advanced teacher-student model update mechanisms is imperative, particularly the dynamic adjustment of EMA's weight parameters to ensure optimal adaptation to the training requirements of diverse stages.